\pdfoutput=1
\documentclass{article} 
\usepackage{collas2026_conference,times}
\usepackage{easyReview}

\usepackage{booktabs}
\usepackage{wrapfig}
\usepackage{algorithm}
\usepackage{algpseudocode}

\usepackage{amsmath,amsfonts,bm}

\def\eqref#1{equation~\ref{#1}}

\def\1{\bm{1}}

\DeclareMathAlphabet{\mathsfit}{\encodingdefault}{\sfdefault}{m}{sl}
\SetMathAlphabet{\mathsfit}{bold}{\encodingdefault}{\sfdefault}{bx}{n}

\usepackage{hyperref}
\hypersetup{
    colorlinks=true,
    linkcolor=red,
    filecolor=magenta,
    urlcolor=blue,
    citecolor=purple,
    pdftitle={Overleaf Example},
    pdfpagemode=FullScreen,
    }

\title{ComMem: Complementary Memory Systems for Test-Time Adaptation of Vision-Language Models}

\author{Guanglong Sun$^{1,6}$ \\
\And 
Shuang Cui$^{2,6}$  \\
\And 
Bo Lei$^{3,6}$ \\
\And
Liyuan Wang$^{4}$ \\
\AND
Zihan Zhai$^{1}$ \\
\And
Hongwei Yan$^{1}$ \\
\And
Hang Su$^{5}$ \\
\And
Jun Zhu$^{5,*}$ \\
\And
Yi Zhong$^{1,*}$ \\
\And
 \\
$^{1}$School of Life Sciences, IDG/McGovern Institute for Brain Research, Tsinghua University, Beijing, China\\
$^{2}$Institute of Software Chinese Academy of Sciences, Beijing, China \\
$^{3}$Beijing Academy of Artificial Intelligence, Beijing, China \\
$^{4}$Department of Psychological and Cognitive Sciences, Tsinghua University, Beijing, China \\
$^{5}$Dept. of Comp. Sci. and Tech., Institute for AI, Tsinghua-Bosch Joint ML Center, \\
  THBI Lab, BNRist Center, Tsinghua University, Beijing, China \\
$^{6}$These authors contributed equally \\
$^{*}$Correspondence:dcszj@tsinghua.edu.cn, zhongyithu@tsinghua.edu.cn \\
}

\collasfinalcopy 

\begin{document}

\maketitle

\begin{abstract}
Test-time adaptation (TTA) of vision-language models (VLMs) is essential for their robust deployment in dynamic, real-world environments. However, existing TTA methods often adapt locally
without accumulating knowledge over time, or operating within a single modality without exploiting VLMs' inherently multi-modal nature. Inspired by the \textbf{Com}plementary \textbf{Mem}ory systems of the biological brain, we propose \textbf{ComMem}, an innovative approach that mimics the distinct but cooperative roles of the hippocampus and neocortex to enable effective TTA for VLMs. ComMem consists of two key components: a fast-adapting detailed memory, akin to the hippocampus, that forms a dynamic visual cache from high-confidence test samples; and a slow-integrating abstract memory, akin to the neocortex, that continually refines global textual prototypes. For each test instance, ComMem jointly optimizes both memory systems to ensure cross-modal consistency. Extensive experiments on 15 benchmark datasets show that ComMem significantly outperforms state-of-the-art methods under both natural distribution shifts and cross-dataset generalization, offering a promising direction for enhancing VLMs' practical adaptability.
\end{abstract}

\section{Introduction}\label{sec:intro}

The advent of pioneering vision-language models (VLMs), such as CLIP~\citep{radford2021learning}, has transformed computer vision by enabling strong zero-shot generalization through learning from vast image-text corpora. Despite their impressive capabilities, these models often suffer significant performance degradation when deployed in real-world environments where the test-time data distribution diverges from their pre-training data distribution~\citep{agarwal2021evaluating, menon2024task}. To address this challenge, test-time adaptation (TTA) has emerged as a powerful paradigm, allowing pre-trained models to adapt to different target domains using only a stream of unlabeled test samples~\citep{liang2025comprehensive, dong2025adapting}.

\begin{figure*}[t]
    \centering
    \includegraphics[width=0.75\textwidth]{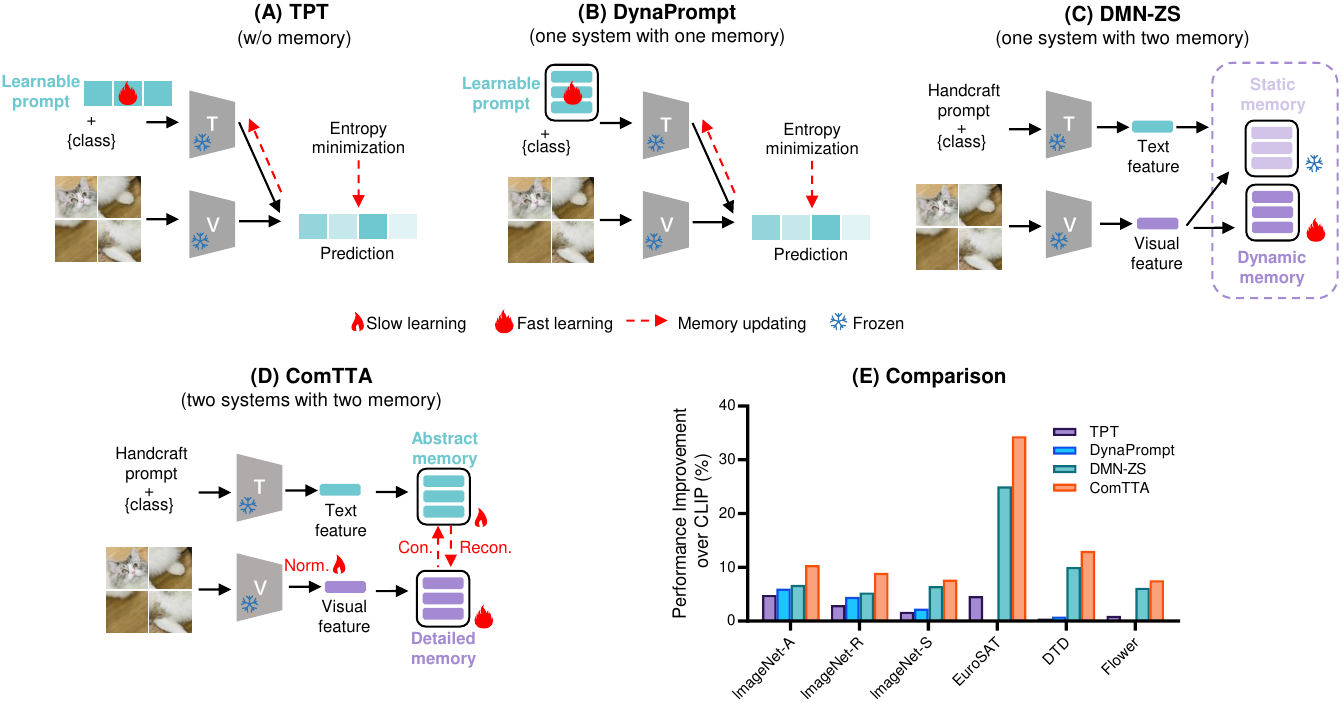}
    \vspace{-0.3cm}
\caption{Comparison of ComMem with recent TTA methods. Norm., Normalization; Con., Consolidation; and Recon., Reconsolidation.}
\label{fig:summaryTTA}
\end{figure*}

Recent TTA efforts for VLMs have evolved from memoryless schemes, in which models are optimized independently for each sample or mini-batch, to more advanced memory-based approaches\footnote{Due to the page limit, we present a more comprehensive summary of related work in Appendix~\ref{sec:related}.}. Early methods, exemplified by TPT~\citep{shu2022test} and its variants such as DiffTPT~\citep{feng2023diverse} and SwapPrompt~\citep{ma2023swapprompt}, operate in an ``amnesic'' manner by adapting to each test instance in isolation without retaining experience over time (Fig.~\ref{fig:summaryTTA}A). Recognizing this limitation, subsequent studies introduced memory mechanisms, which can be broadly categorized by their adaptive component: some accumulate knowledge at the textual level through prompt tuning, such as HisTPT~\citep{zhang2024historical} and DynaPrompt~\citep{xiao2025dynaprompt} (Fig.~\ref{fig:summaryTTA}B), whereas others maintain dynamic caches at the visual representation level, such as TDA~\citep{karmanov2024efficient} and DMN-ZS~\citep{zhang2024dual} (Fig.~\ref{fig:summaryTTA}C). However, restricting adaptation to single modality limits the model's ability to align visual and textual representations effectively. Without coordinated cross-modal updates, these methods fail to balance the complementary fast (plastic) and slow (stable) learning dynamics, thereby limiting the ability to rapidly adapt to new domain-specific cues while preserving the robustness of pre-trained knowledge under distribution shifts (Fig.~\ref{fig:summaryTTA}D-E).


\begin{figure*}[t]
    \centering
    \includegraphics[width=0.95\textwidth]{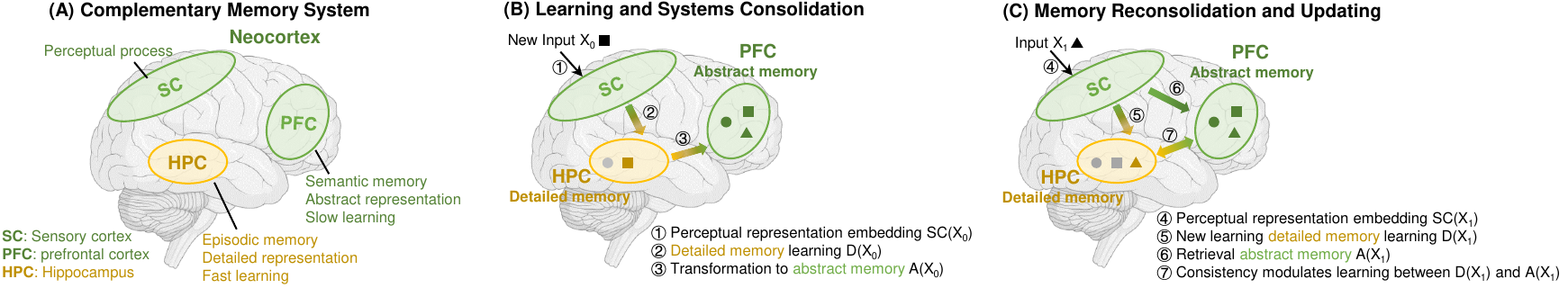}
\caption{Overview of the complementary memory systems theory~\citep{mcclelland1995there, tonegawa2018role, lei2025reconstructing} in neuroscience. 
}
\vspace{-0.5cm}
\label{fig:ComplementaryMemory}
\end{figure*}

In neuroscience, \emph{complementary memory systems} theory~\citep{mcclelland1995there, tonegawa2018role, lei2025reconstructing}, also known as \emph{complementary learning systems}, has been widely acknowledged to achieve an integration of new experience into prior knowledge for optimal behavioral output.
This theory posits a functional synergy between two distinct memory systems (Fig.~\ref{fig:ComplementaryMemory}A): the hippocampus (HPC) rapidly encodes episodic and detailed memories; the neocortex (NC) slowly forms semantic and generalizable knowledge. Over time, memory traces are gradually transferred from the HPC to the NC through \emph{systems consolidation} (Fig.~\ref{fig:ComplementaryMemory}B)~\citep{golbabaei2025post} process. 
More importantly, a recent work revealed that recalled of an NC-dependent memory can induce \emph{memory reconsolidation} phase in HPC (Fig.~\ref{fig:ComplementaryMemory}C), allowing the HPC to incorporate new information and reconstruct an memory trace~\citep{lei2025reconstructing}.


Motivated by this well-established biological mechanism, we introduce \textbf{ComMem}, an innovative TTA approach that instantiates the dynamics of complementary memory systems for robust VLMs adaptation (Fig.~\ref{fig:ComMem}, see pseudo-code in Appendix Alg.~\ref{alg:commem}). ComMem models the fast, plastic learning of the HPC as a detailed memory cache that rapidly stores and updates visual features from high-confidence test samples. In parallel, it models the NC's slow, stable learning as an abstract memory composed of global textual prototypes, which evolve gradually through a process analogous to memory consolidation. For each incoming test sample, ComMem simulates memory reconsolidation by retrieving from both memory systems and jointly optimizing them via learnable residuals to enforce cross-modal consistency. This enables the model not only to adapt to individual samples but also to progressively accumulate and refine domain knowledge across the test data stream.

We evaluate ComMem with extensive experiments across 15 challenging benchmark datasets. Our results demonstrate that ComMem consistently and significantly outperforms current state-of-the-art methods in both robustness to natural distribution shifts (e.g., ImageNet-A/R/S) and broad cross-dataset generalization scenarios. ComMem achieves a new state-of-the-art average accuracy of 48.84\% on ImageNet-OOD variants with ResNet-50~\citep{resnet} and 65.36\% with ViT-B/16~\citep{vit}. Furthermore, our approach ensures both strong performance and efficiency, incurring substantially less computational overhead than prominent prompt-tuning methods like TPT~\citep{shu2022test}. Detailed ablation studies confirm that the synergistic combination of the fast-adapting detailed memory and the slow-integrating abstract memory, alongside our reconsolidation process, is critical to its success.

Our main contributions are as follows:

\begin{itemize}
    \item We introduce ComMem, a brain-inspired approach to TTA of VLMs that, for the first time, operationalizes the distinct fast and slow learning dynamics underlying the brain's complementary memory systems.
    \item We design a synergistic mechanism involving a fast-updating detailed cache and a slow-consolidating abstract memory, which are jointly optimized through a reconsolidation process for cross-modal consistency.
    \item We conduct extensive experiments on 15 benchmark datasets, demonstrating that ComMem significantly outperforms state-of-the-art TTA methods under distribution shifts and cross-dataset generalization.
\end{itemize}

\section{Preliminaries}\label{prelim}

In this section, we describe the problem formulation of zero-shot VLMs classification, test-time adaptation in VLMs, and our brain-inspired motivations.

\subsection{Formulation}

\paragraph{Zero-Shot VLMs Classification}

Representative VLMs such as CLIP~\citep{radford2021learning} consist of a visual encoder $\mathcal{E}_v(\cdot)$ and a textual encoder $\mathcal{E}_t(\cdot)$ that project images and text into a shared, high-dimensional embedding space $\mathbb{R}^D$. For a $C$-way classification task, the class names $\{y_c\}_{c=1}^C$ are embedded into textual features using prompt ensembling, e.g., ``a photo of a \{CLASS\}'', yielding a text prototype matrix $\mathbf{P}^t \in \mathbb{R}^{D \times C}$. 

For an input image $\mathbf{x}$, its visual feature $\mathbf{f}_v = \mathcal{E}_v(\mathbf{x})$ is compared with the text features using cosine similarity, producing the zero-shot class posterior:
\begin{equation}\label{clip_classifier}
p(y=c|\mathbf{x}) = \frac{\exp(\text{sim}(\mathbf{f}_v, \mathbf{p}^t_c) / \tau)}{\sum_{j=1}^C \exp(\text{sim}(\mathbf{f}_v, \mathbf{p}^t_j) / \tau)},
\end{equation}
where $\mathbf{p}^t_c$ is the $c$-th text prototype, $\text{sim}(\cdot,\cdot)$ denotes cosine similarity, and $\tau$ is the temperature hyperparameter. 

\paragraph{Test-Time Adaptation in VLMs}\label{sec:tta}

We consider an \textit{online streaming} test-time adaptation (TTA) scenario, where a pre-trained VLM $(\mathcal{E}_v,\mathcal{E}_t)$ is deployed in a target domain with unlabeled test samples $\{\mathbf{x}_s\}_{s=1}^S$ arriving sequentially. 
No ground-truth labels are available, and the target distribution $\mathcal{P}_\text{tgt}(x)$ differs from the source distribution $\mathcal{P}_\text{src}(x)$. 
The objective is to adapt the model \textit{on-the-fly} to minimize the expected target error:
\begin{equation}
\min_{\theta} \ \mathbb{E}_{\mathbf{x} \sim \mathcal{P}_\text{tgt}} \big[ \ell(\hat{y}(\mathbf{x}; \theta), y) \big],
\end{equation}
where $\theta$ denotes the parameters of the model and its auxiliary adaptation modules, $\ell(\cdot)$ is the classification loss, and $\hat{y}$ represents the model's prediction. 

Unlike regular domain adaptation, TTA operates without access to labeled target data. 
In the \textit{online streaming} setting, test samples arrive sequentially in a continuous stream. Accordingly, the adaptation algorithm must satisfy the following dynamics:
\begin{align}
\mathbf{x}_s &\sim \mathcal{P}_\text{tgt}, \\
\theta_{s} &= \text{Update}(\theta_{s-1}, \mathbf{x}_s), \\
\hat{y}_s &= f(\mathbf{x}_s; \theta_s),
\end{align}
where $s$ indexes the time step, $\text{Update}(\cdot)$ denotes an unsupervised adaptation rule based on model confidence, entropy, or consistency.

A central challenge lies in \textit{rapidly adapting} to domain-specific variations to mitigate error accumulation from self-training noise, while simultaneously \textit{maintaining stability} to prevent catastrophic forgetting. 
To address this trade-off, we draw inspirations from the robust biological brain, especially the \textit{complementary memory systems} theory.

\subsection{Brain-Inspired Motivation}

\begin{figure}[t]
    \centering
    \includegraphics[width=0.50\textwidth]{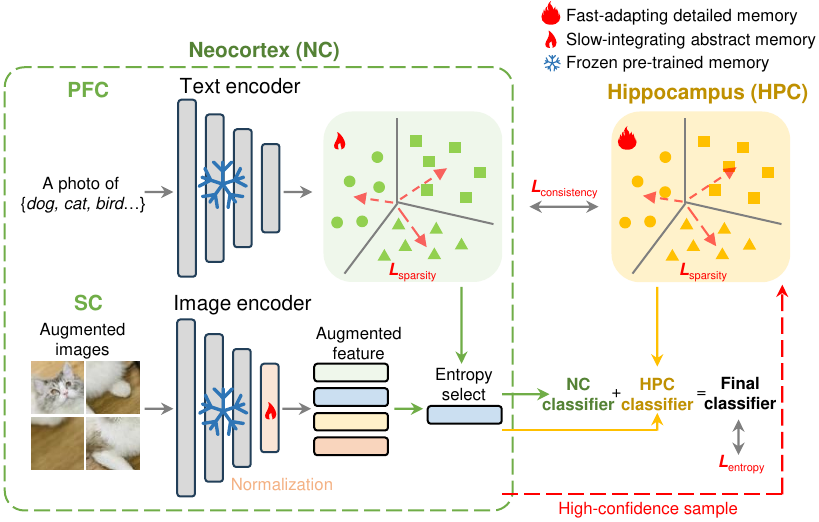}
\caption{The proposed ComMem framework for test-time adaptation of vision-language models. 
}
\label{fig:ComMem}
\end{figure}

The remarkable adaptability of the biological brain arises from the interaction between the following two memory systems (Fig.~\ref{fig:ComplementaryMemory}A)~\citep{mcclelland1995there, tonegawa2018role, lei2025reconstructing}: 
\begin{itemize}
    \item \textbf{Neocortex (NC):} The NC encodes semantic, abstract knowledge (e.g., the concept of a ``dog'') through \textit{slow and integrative learning}, ensuring knowledge stability and generalization across experiences.
    \item \textbf{Hippocampus (HPC):} The HPC rapidly captures specific, episodic experiences (e.g., ``the particular dog I saw in the park''), supporting \textit{fast, instance-based learning} with high plasticity.
\end{itemize}

The key mechanisms underlying interactions of the two memory systems are \textit{systems consolidation} and \textit{systems reconsolidation}.  
During \textit{systems consolidation} (Fig.~\ref{fig:ComplementaryMemory}B), episodic traces encoded in the HPC are gradually integrated into cortical representations within the NC, enabling the abstraction of long-term generalizable knowledge.  
Conversely, during \textit{systems reconsolidation} (Fig.~\ref{fig:ComplementaryMemory}C), when novel stimuli are encountered, the NC retrieves existing abstract representation memory to interpret new episodic inputs from the HPC. Both memory traces are then jointly updated, allowing the brain to maintain coherent and adaptive representations over time.

Motivated by this effective synergy between fast and slow learning systems, our proposed \textbf{ComMem} framework (Fig.~\ref{fig:ComMem}) computationally instantiates these dynamics for TTA in VLMs.  
Specifically, we model a \textbf{detailed memory cache}, analogous to the HPC, that rapidly stores and updates high-confidence visual features from the test data stream to enable quick adaptation; and an \textbf{abstract memory systems}, analogous to the NC, represented by slowly evolving textual prototypes that accumulate stable, generalizable knowledge.  
For each incoming test sample, ComMem performs a computational analogue of memory consolidation and memory reconsolidation, where information from both memories is retrieved and jointly optimized to ensure cross-modal alignment, followed by gradual consolidation of reliable updates into the abstract memory.

\section{Method}\label{sec:method}
The ComMem framework is illustrated in Fig.~\ref{fig:ComMem} (see pseudo-code in Appendix Alg.~\ref{alg:commem}). We first introduce the two complementary memory components and their distinct update mechanisms, then detail the joint optimization process that computationally simulates complementary memory systems.

\subsection{Visual Representation Encoding}\label{sec:entropy_encoding}

To facilitate rapid, instance-based adaptation, ComMem maintains an episodic memory cache $\mathcal{M}$ that stores class-conditional visual instances as feature–entropy pairs $\{(\mathbf{f}_i, \mathcal{H}_i)\}_{i=1}^{|\mathcal{M}_c|}$, where each class $c$ retains up to $K$ entries ranked by confidence (i.e., inverse entropy). 
This cache serves as the HPC-like memory, corresponding to precise, high-certainty visual representations from the target domain.

\paragraph{Entropy-Guided Encoding.}
For each test image $\mathbf{x}_t$, we generate a set of $N$ randomly augmented views $\{\mathbf{x}_t^{(i)}\}_{i=1}^N$ and extract their visual embeddings $\mathbf{f}_v^{(i)} = \mathcal{E}_v(\mathbf{x}_t^{(i)})$ using the frozen visual encoder $\mathcal{E}_v$. 
Each embedding is passed through the classifier (Eq.~\ref{clip_classifier}) to obtain the predicted probability distribution 
$p^{(i)} = p(y|\mathbf{f}_v^{(i)})$. 
The prediction entropy $\mathcal{H}^{(i)}$ of each view is then computed as:
\begin{equation}
\mathcal{H}^{(i)} = - \sum_{c=1}^{C} p^{(i)}_c \log p^{(i)}_c,
\label{eq:entropy}
\end{equation}
where $C$ is the number of classes and lower entropy indicates higher prediction confidence. 

To ensure that only reliable representations contribute to adaptation, we perform a confidence-weighted aggregation of these augmented views:
\begin{equation}
w^{(i)} = \frac{\exp(-\mathcal{H}^{(i)} / \tau_e)}{\sum_{j=1}^N \exp(-\mathcal{H}^{(j)} / \tau_e)}, 
\quad 
\mathbf{f}_v^* = \sum_{i=1}^N w^{(i)} \mathbf{f}_v^{(i)},
\label{eq:entropy_weight}
\end{equation}
where $\tau_e$ is an entropy temperature controlling the sharpness of the weighting distribution. 
The resulting $\mathbf{f}_v^*$ represents an entropy-refined, confidence-weighted embedding of the target instance, while its pseudo-label $\hat{y} = \arg\max p(\cdot|\mathbf{f}_v^*)$ is subsequently used to update the detailed memory cache.

\subsection{HPC Detailed Memory Fast Adaptation}
\label{sec:hpc_update}

For an incoming feature $\mathbf{f}_v^*$ with pseudo-label $\hat{y}$ and entropy $\mathcal{H}$, the episodic cache $\mathcal{M}$ is updated as follows:
\begin{itemize}
    \item \textbf{If the cache is not full ($|\mathcal{M}_{\hat{y}}| < K$):}  
    The new feature–entropy pair $(\mathbf{f}_v^*, \mathcal{H})$ is directly added to the cache of class $\hat{y}$.
    \item \textbf{If the cache is full ($|\mathcal{M}_{\hat{y}}| = K$):}  
    The cache entries for class $\hat{y}$ are sorted by their entropy. Let $(\mathbf{f}_{\max}, \mathcal{H}_{\max})$ denote the entry with the highest entropy (i.e., lowest confidence).  
    If the new feature is less confident ($\mathcal{H} > \mathcal{H}_{\max}$), the cache remains unchanged. Otherwise, it replaces this entry through an \textit{instance-level evolution} step.
\end{itemize}

\paragraph{HPC Memory Update Rule.}\label{sec:hpc_update}
Before replacing the least confident entry $(\mathbf{f}_{\max}, \mathcal{H}_{\max})$, the incoming high-confidence feature $\mathbf{f}_v^*$ is evolved to form a more robust representation.  
In this step, only the pair with $\mathcal{H}_{\max}$ is updated, and the prototype of its pseudo-label class is computed as 
$\mathbf{p}^v_{\hat{y}} = \text{mean}(\{\mathbf{f}_i \in \mathcal{M}_{\hat{y}}\})$.  
The final stored representation $\mathbf{f}'_v$ is obtained via a weighted fusion of the class prototype and the new instance:
\begin{equation} \label{eq:hpc_evo}
\mathbf{f}'_v = (1-\lambda_{\text{hpc}})\mathbf{p}^v_{\hat{y}} + \lambda_{\text{hpc}}\mathbf{f}_v^*,
\end{equation}
where $\lambda_{\text{hpc}}$ controls the update plasticity.  
A larger $\lambda_{\text{hpc}}$ 
corresponds to a \textit{fast, instance-dominant update}, while a smaller value leads to a \textit{slow, prototype-dominant integration}.  
The updated pair $(\mathbf{f}'_v, \mathcal{H})$ then replaces $(\mathbf{f}_{\max}, \mathcal{H}_{\max})$ in $\mathcal{M}_{\hat{y}}$.

\paragraph{Detailed Memory Classifier Generation.}
After each update, the complete set of visual prototypes 
$\mathbf{P}^v = [\mathbf{p}^v_1, \dots, \mathbf{p}^v_{C_{\text{active}}}] \in \mathbb{R}^{D \times C_{\text{active}}}$  
is re-computed by averaging all features currently stored in each active class.  
This matrix $\mathbf{P}^v$ serves as the base classifier for the detailed (HPC-like) memory and is subsequently refined by a learnable residual term $\delta_v$ during the test-time optimization stage (see Sec.~\ref{sec:optimization}).

\subsection{NC Abstract Memory Slow Integration}\label{sec:nc_evolution}

To model the stable, semantic knowledge of the neocortex, the \textit{Abstract Memory} is represented by a set of global textual prototypes $\mathbf{P}^t \in \mathbb{R}^{D \times C}$.  
This component reflects the model’s pre-trained, generalized knowledge and undergoes a \textit{slow integration} process during test time.

\paragraph{NC Memory Update Rule.}\label{sec:nc_update}
This gradual update mechanism integrates newly acquired task-specific knowledge back into the stable abstract memory, analogous to systems consolidation in neuroscience.  
The update is triggered \textit{after} the test-time optimization step (Sec.~\ref{sec:optimization}) for each test instance $\mathbf{x}_t$.  
If the final prediction is sufficiently confident (i.e., prediction entropy is below a threshold $\tau_{\text{conf}}$), the locally optimized prototype $\hat{\mathbf{P}}^t_{\mathrm{local}}$ (defined in Eq.~\ref{eq:local_text_proto}) is used to update the global abstract memory $\mathbf{P}^t$. The update follows a weighted fusion controlled by the rate $\lambda_{\text{nc}}$:
\begin{equation} \label{eq:nc_evo}
\mathbf{P}^t \leftarrow (1-\lambda_{\text{nc}})\mathbf{P}^t + \lambda_{\text{nc}}\hat{\mathbf{P}}_{\mathrm{local}}^{t},
\end{equation}
where $\lambda_{\text{nc}}$ is the update rate. A value approaching 0
results in a \textit{slow, stable integration} that preserves the stability of pre-trained knowledge (an NC trait). Conversely, a value approaching 1
leads to a \textit{fast, unstable update}, risking catastrophic forgetting. This ensures $\mathbf{P}^t$ evolves gradually while slowly absorbing new, high-confidence information from the target domain.

\subsection{Test-Time Optimization}\label{sec:optimization}

For each test sample, ComMem retrieves and integrates both memory systems, performing a single-step optimization over a small set of learnable parameters to achieve adaptive yet stable prediction. These parameters correspond to distinct brain-inspired modules with fast and slow learning rates:

\begin{itemize}
    \item \textbf{Visual Residuals ($\delta_v \in \mathbb{R}^{D \times C_{\text{active}}}$):}  
    A learnable residual term added to the visual prototypes $\mathbf{P}^v$ (retrieved from the hippocampal-like cache) to construct a sample-specific visual classifier:
    \begin{equation}
    \hat{\mathbf{P}}^v = \text{Normalize}(\mathbf{P}^v + \delta_v).
    \end{equation}
    This component belongs to the fast-learning HPC system, responsible for adapting episodic visual details, and is updated with a learning rate $lr_v$.

    \item \textbf{Textual Residuals ($\delta_t \in \mathbb{R}^{D \times C}$):}  
    A learnable residual added to the global textual prototypes $\mathbf{P}^t$ to yield a context-adaptive local classifier:
    \begin{equation} \label{eq:local_text_proto}
    \hat{\mathbf{P}}^t_{\text{local}} = \text{Normalize}(\mathbf{P}^t + \delta_t).
    \end{equation}
    This component is associated with the slow-learning NC system, which encodes abstract, semantic knowledge, and is updated with a learning rate $lr_t$.

    \item \textbf{Normalization Layers:}  
    The parameters of layer normalization (LN) or batch normalization (BN) layers, denoted $\theta_n$, within the visual encoder are fine-tuned to enhance low-level feature adaptation:
    \begin{equation}
    \hat{\mathbf{f}}_v = \text{LN}_{\theta_n}(\mathbf{f}_v) \ \text{or} \ \text{BN}_{\theta_n}(\mathbf{f}_v).
    \end{equation}
    These parameters belong to the NC system, capturing slower cortical adaptation of perceptual representations, and are updated with a learning rate $lr_n$.
\end{itemize}

\paragraph{Prediction Integration.}
Given a visual feature $\mathbf{f}_v$, ComMem integrates predictions from both the abstract (NC) and detailed (HPC) systems:
\begin{equation}
\mathbf{z}_{\text{final}} = \mathbf{f}_v^\top \hat{\mathbf{P}}^t_{\text{local}} + \mathcal{A}(\mathbf{f}_v^\top \hat{\mathbf{P}}^v),
\end{equation}
where $\mathcal{A}(s) = \alpha \exp(-\beta (1 - s))$ is an affinity function that adaptively weights the contribution of the detailed memory, following~\citep{zhang2024dpe}. This integration mimics biological memory reconsolidation, where episodic and semantic traces jointly determine the final perception or decision.

\paragraph{Joint Optimization Objective.}
The learnable parameters $(\delta_t, \delta_v, \theta_n)$ are optimized in a single gradient step by minimizing a composite loss function:
\begin{equation}
\mathcal{L}_{\text{total}} = \mathcal{L}_{\text{ent}} + \lambda_{\text{align}}\mathcal{L}_{\text{align}} + \lambda_{\text{sparse}}\mathcal{L}_{\text{sparse}}.
\end{equation}
Each term is framed with neuro-computational
analogues:

\begin{itemize}
    \item \textbf{Entropy Minimization ($\mathcal{L}_{\text{ent}}$):}  
    Reduces the prediction uncertainty by minimizing the entropy of the final logits, promoting confident, stable decisions:
    \begin{equation}
    \mathcal{L}_{\text{ent}} = \mathcal{H}\!\left(\text{Softmax}(\mathbf{z}_{\text{final}} / \tau)\right).
    \end{equation}
    This mirrors the neural process of uncertainty resolution during memory updating~\citep{mason2017role,radvansky1995uncertainty}.

    \item \textbf{Inter-Memory Consistency ($\mathcal{L}_{\text{align}}$):}  
    Encourages coherence between the visual and textual representations by aligning their respective prototypes via a symmetric InfoNCE loss~\citep{oord2018representation}:
    \begin{equation}
    \begin{split}
    \mathcal{L}_{\text{align}} = -\frac{1}{C_{\text{active}}}\sum_{c=1}^{C_{\text{active}}} \Biggl( & 
    \log\frac{\exp(\hat{\mathbf{p}}_{t,c}^\top \hat{\mathbf{p}}_{v,c} / \tau_c)}{\sum_{j=1}^{C_{\text{active}}}\exp(\hat{\mathbf{p}}_{t,c}^\top \hat{\mathbf{p}}_{v,j} / \tau_c)} \\
    & + \log\frac{\exp(\hat{\mathbf{p}}_{v,c}^\top \hat{\mathbf{p}}_{t,c} / \tau_c)}{\sum_{j=1}^{C_{\text{active}}}\exp(\hat{\mathbf{p}}_{v,c}^\top \hat{\mathbf{p}}_{t,j} / \tau_c)} 
    \Biggr).
    \end{split}
    \end{equation}
    This term maintains consistency between abstract (NC) and detailed (HPC) representations, ensuring coherent reconsolidation~\citep{colyer2025phase}.

    \item \textbf{Intra-Memory Sparse Coding ($\mathcal{L}_{\text{sparse}}$):}  
    Inspired by sparse neural coding~\citep{wixted2014sparse,palm2013neural}, which minimizes representational overlap across memory traces, this term penalizes prototype redundancy within each memory system.  
    For a prototype matrix $\mathbf{P} \in \{\hat{\mathbf{P}}^t_{\text{local}}, \hat{\mathbf{P}}^v\}$:
    \begin{equation}
    \mathcal{L}_{\text{sparse}}(\mathbf{P}) = \frac{1}{C(C-1)}\sum_{i \neq j} (\mathbf{p}_i^\top \mathbf{p}_j)^2.
    \end{equation}
\end{itemize}

The optimization is executed with a single-step update using the AdamW optimizer~\citep{loshchilov2017decoupled}.  
After adaptation, the refined residuals yield the final prediction for the current sample.  
If the prediction confidence surpasses a predefined threshold, the global semantic prototypes $\mathbf{P}^t$ are further updated via Eq.~\ref{eq:nc_evo}, preparing the system for subsequent samples in the test data stream.

\begin{table*}[t]
\renewcommand\arraystretch{1.0}
\renewcommand{\tabcolsep}{2pt}
\vspace{-1.5cm}
\caption{\textbf{Performance comparison under natural distribution shifts.} Top-1 accuracy (\%) of all evaluated methods using ResNet-50 and ViT-B/16 as CLIP vision backbones. The best and second-best results are marked in \textbf{bold} and \underline{underlined}, respectively.}
  \centering
  \small
  \resizebox{\linewidth}{!}{%
  \begin{tabular}{lccccccccc}
    \toprule
    {Method}  &Venue  &Buffer  & ImageNet   &  ImageNet-A  &  ImageNet-V2  & ImageNet-R & ImageNet-S   
    & {Average}  & {OOD Average}     \\
  \midrule
  CLIP-ResNet-50~\citep{radford2021learning} &ICML21   & $\times$    &58.16&	21.83&	51.41	&56.15	&33.37&	44.18&	40.69           \\ 
  \midrule
  Ensemble    & - & $\times$      &  59.81&	23.24&	52.91	&{60.72}	&35.48&	46.43&	43.09  \\
  CoOp~\citep{zhou2022learning}   &IJCV22  & $\times$      &  63.33 &  23.06  &   55.40     &  56.60   &  34.67   &     46.61   &     42.43       \\
  \midrule
  TPT~\citep{shu2022test}    &NeurIPS22  &  $\times$      &  60.74 & 26.67  &    54.70   &  59.11   &  35.09   &    47.26    &     43.89    \\
  DiffTPT~\citep{feng2023diverse} &ICCV23  & $\times$  & 60.80 & \underline{31.06} & 55.80 & 58.80 & 37.10 & 48.71 & 45.69 \\
  DynaPrompt~\citep{xiao2025dynaprompt} &ICLR25 &$\checkmark$ &61.56	&27.84 &55.12	&60.63 &35.64			&48.16	&44.81  \\
  TDA~\citep{karmanov2024efficient} &CVPR24 &$\checkmark$  & 61.35 & 30.29 & 55.54 & 62.58 & 38.12 & 49.58 & 46.63 \\ 
  TPS~\citep{sui2025just}	&WACV25 &$\checkmark$ &61.47&	30.48&	54.96&	62.87&	37.14&	49.38&	46.36 \\
  DPE~\citep{zhang2024dpe} &NeurIPS24 &$\checkmark$  & 63.41 & 30.15 & \underline{56.72} & \underline{63.72} & \underline{40.03} &\underline{50.81} &\underline{47.66} \\
  DMN-ZS~\citep{zhang2024dual} &CVPR24 &$\checkmark$	&\underline{63.87} &	28.57&	56.12&	61.44&	39.84&	49.97&	46.49\\
  \textbf{ComMem} &\textbf{Ours} &$\checkmark$ &\textbf{64.05} & \textbf{32.21} & \textbf{56.95} & \textbf{65.10}  & \textbf{41.09} &\textbf{51.88} & \textbf{48.84}  \\
  \midrule
  \midrule
  CLIP-ViT-B/16~\citep{radford2021learning} &ICML21   & $\times$     & 66.73&	47.87&	60.86&	73.98&	46.09&	59.11&	57.20             \\
  \midrule
  Ensemble    & - & $\times$       &  68.34&	49.89&	61.88&	77.65&	48.24&	61.20&	59.42  \\
  CoOp~\citep{zhou2022learning}    &IJCV22  & $\times$      &  71.51 &  49.71  &   64.20     &  75.21   &  47.99   &   61.72     &   59.28  \\
  \midrule
  TPT~\citep{shu2022test}    &NeurIPS22  &  $\times$       &  68.98  & 54.77  & 63.45       &  77.06   &  47.94   &   62.44     &   60.81 \\
  DiffTPT~\citep{feng2023diverse}    &ICCV23  & $\times$       &  70.30  & 55.68  & 65.10       &  75.00   &  46.80   &   62.28     &   60.52 \\
  DynaPrompt~\citep{xiao2025dynaprompt} &ICLR25 &$\checkmark$ &69.61  &56.17  &64.67  &78.17  &48.22  &63.37  &61.81  \\
  TDA~\citep{karmanov2024efficient} &CVPR24 & $\checkmark$ & 69.51 & 60.11 & 64.67 & 80.24 & 50.54 & 65.01 & 63.89 \\ 
  TPS~\citep{sui2025just}	&WACV25 & $\checkmark$ &70.19&	60.08&	64.73&	80.27&	49.95&	65.04&	63.76 \\
  DPE~\citep{zhang2024dpe} &NeurIPS24 & $\checkmark$ & 71.91 & 59.63 & \textbf{65.44} & \underline{80.40} & 52.26 & \underline{65.93} & \underline{64.43} \\
  DMN-ZS~\citep{zhang2024dual}	 &CVPR24 & $\checkmark$ &\underline{72.25}&	58.28&	65.17&	78.55&	\underline{53.20} &	65.49&	63.80\\
  \textbf{ComMem} &\textbf{Ours} & $\checkmark$ &\textbf{72.16} & \textbf{61.32}  & \underline{65.38} & \textbf{81.52} & \textbf{53.21} &\textbf{66.72} & \textbf{65.36}   \\
    \bottomrule
  \end{tabular}}
\label{tab:ood-main}
\end{table*}

\section{Experiments}

In this section, we describe the experimental setups (further detailed in Appendix Sec.~\ref{sec:setup}), including datasets, implementation details and baselines, and then present the experimental results with an in-depth analysis.


\subsection{Performance Evaluation}

\paragraph{Robustness to Natural Distribution Shifts.}
In Table~\ref{tab:ood-main}, we present the performance comparison on ImageNet and its variants. Our proposed ComMem consistently achieves state-of-the-art performance across both CLIP~\citep{radford2021learning} backbones. With the ResNet-50~\citep{resnet} backbone, ComMem attains an average Out-of-Distribution (OOD) accuracy of 48.84\%, surpassing the next-best method, DPE~\citep{zhang2024dpe}, which scores 47.66\%. A similar trend is observed with the ViT-B/16~\citep{vit} backbone, where ComMem achieves an average OOD accuracy of 65.36\%, outperforming all other competing methods. These results underscore ComMem's superior ability to generalize to unseen domains throughout training. 

\begin{table*}[th]
\renewcommand\arraystretch{1.0}
\renewcommand{\tabcolsep}{2.0pt}
\vspace{-1.5cm}
  \caption{\textbf{Performance comparisons on cross-dataset generalization.} Top-1 accuracy (\%) of all evaluated methods using two CLIP vision backbones. The best and second-best results are marked in \textbf{bold} and \underline{underlined}, respectively. Results for DynaPrompt~\citep{xiao2025dynaprompt} on Cars~\citep{krause20133d} and SUN397~\citep{xiao2010sun} are marked as “–” due to out-of-memory errors during reproduction.}
  \centering
  \tiny
  \resizebox{\linewidth}{!}{
  \begin{tabular}{lccccccccccc}
      \toprule
      Method & Aircraft & Caltech & Cars & DTD & EuroSAT & Flower & Food101 & Pets & SUN397 & UCF101 & Average \\
      \midrule
      CLIP-ResNet-50~\citep{radford2021learning} & 15.66 & 85.88 & 55.70 & 40.37 & 23.69 & 61.75 & 73.97 & 83.57 & 58.80 & 58.84 & 55.82 \\
      \midrule
      Ensemble & 16.11 & 87.26 & 55.89 & 40.37 & 25.79 & 62.77 & 74.82 & 82.97 & 60.85 & 59.48 & 56.63 \\
      CoOp~\citep{zhou2022learning} & 15.12 & 86.53 & 55.32 & 37.29 & 26.20 & 61.55 & 75.59 & 87.00 & 58.15 & 59.05 & 56.18 \\
      \midrule
      TPT~\citep{shu2022test}  & 17.58 & 87.02 & 58.46 & 40.84 & 28.33 & 62.69 & 74.88 & 84.49 & 61.46 & 60.82 & 57.66 \\
      DiffTPT~\citep{feng2023diverse} & 17.60 & 86.89 & \textbf{60.71} & 40.72 & 41.04 & 63.53 & \textbf{79.21} & 83.40 & 62.72 & 62.67 & 59.85 \\
      DynaPrompt~\citep{xiao2025dynaprompt}  &16.44	&88.07	&-	&41.19	&23.41	&61.84	&75.73	&83.07	&-	&59.87	&- \\
      TDA~\citep{karmanov2024efficient}  & 17.61 & 89.70 & 57.78 & 43.74 & 42.11 & 68.74 & 77.75 & 86.18 & 62.53 & \underline{64.18}  & 61.03\\ 
      DPE~\citep{zhang2024dpe}  & 19.80 & \underline{90.83} & 59.26 & \underline{50.18} & 41.67 & 67.60 & 77.83 & 85.97 & 64.23 & 61.98 & 61.93 \\
      DMN-ZS~\citep{zhang2024dual}  & \textbf{22.77} & 90.14 & 60.02 & 50.41 & \underline{48.75} & \underline{67.93} & 76.70 & \textbf{86.78} & \underline{64.39} & \textbf{65.34} & \underline{63.32} \\
      \textbf{ComMem (Ours)} & \underline{21.33} & \textbf{91.32} &\underline{60.19} &\textbf{53.43} &\textbf{58.05} &\textbf{69.31} &\underline{78.04} &\underline{86.59} &\textbf{65.11} &62.62 &\textbf{64.60} \\
      \midrule
      \midrule
      CLIP-ViT-B/16~\citep{radford2021learning} & 23.67 & 93.35 & 65.48 & 44.27 & 42.01 & 67.44 & 83.65 & 88.25 & 62.59 & 65.13 & 63.58 \\
      \midrule
      Ensemble & 23.22 & 93.55 & 66.11 & 45.04 & 50.42 & 66.99 & 82.86 & 86.92 & 65.63 & 65.16 & 64.59 \\
      CoOp~\citep{zhou2022learning} & 18.47 & 93.70 & 64.51 & 41.92 & 46.39 & 68.71 & 85.30 & 89.14 & 64.15 & 66.55 & 63.88 \\
      \midrule
      TPT~\citep{shu2022test}  & 24.78 & 94.16 & 66.87 & 47.75 & 42.44 & 68.98 & 84.67 & 87.79 & 65.50 & 68.04 & 65.10 \\
      DiffTPT~\citep{feng2023diverse} & 25.60 & 92.49 & 67.01 & 47.00 & 43.13 & 70.10 & \textbf{87.23} & 88.22 & 65.74 & 62.67 &65.47 \\
      DynaPrompt~\citep{xiao2025dynaprompt} &24.33	&94.32	&67.65	&47.96	&42.28	&69.95	&85.42	&88.28	&66.32	&68.72	&65.52 \\
      TDA~\citep{karmanov2024efficient} & 23.91 & 94.24 & 67.28 & 47.40 & 58.00 & 71.42 & 86.14 & 88.63 & 67.62 & 70.66 & 67.53 \\ 
      TPS~\citep{sui2025just} & 26.34 & 95.05 & \textbf{69.09} & 50.47 & 44.48 & 71.54 & 85.23 & 87.35 & 69.98 & 71.00 & 66.96 \\
      DPE~\citep{zhang2024dpe} & 28.95 & 94.81 & 67.31 & 54.20 & 55.79 & \underline{75.07} & 86.17 & 91.14 & 70.07 & 70.44 & 69.40 \\
      DMN-ZS~\citep{zhang2024dual}  & \underline{30.03} & \textbf{95.38} & \underline{67.96} & \textbf{55.85} & \underline{59.43} & 74.49 & 85.08 & \underline{92.04} & \underline{70.18} & \textbf{72.51} & \underline{70.30}  \\
      \textbf{ComMem (Ours)} & \textbf{30.45} &\underline{95.21} &67.89 &\underline{55.50} &\textbf{61.80} &\textbf{75.56} &\underline{86.48} &\textbf{92.29} &\textbf{70.48} &\underline{71.74} & \textbf{70.74} \\
      \bottomrule
    \end{tabular}
  } 
  \label{tab:cross}
\end{table*}

\paragraph{Cross-Dataset Generalization.}
We further assess the generalization of ComMem on 10 diverse datasets, with results summarized in Table~\ref{tab:cross}. The significant distributional differences across these datasets pose a substantial challenge for adaptation methods. Nevertheless, ComMem demonstrates remarkable robustness. With the ResNet-50~\citep{resnet} backbone, ComMem again leads with an average accuracy of 64.60\%, outperforming the strong baselines DMN-ZS~\citep{zhang2024dual} of 63.32\% and DPE~\citep{zhang2024dpe} of 61.93\%. Similarly, with the ViT-B/16~\citep{vit} backbone, ComMem achieves the highest average accuracy of 70.74\%. The consistent performance gains highlight the adaptability and effectiveness of our brain-inspired framework.

\subsection{Ablation Studies}
\label{sec:ablation}

\begin{figure*}[t]
    \centering
    \includegraphics[width=1.0\textwidth]{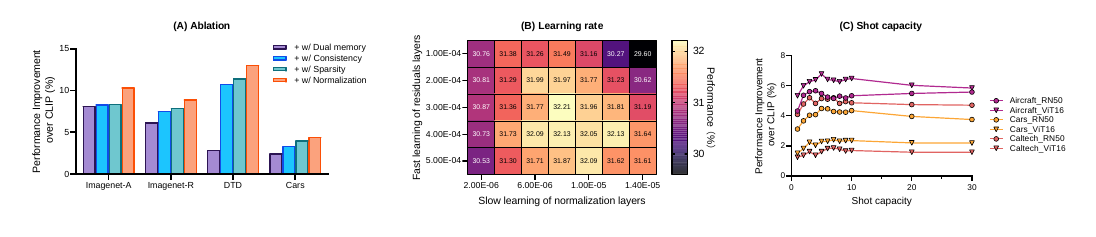}
     \vspace{-1cm}
    \caption{
    \textbf{Ablation and hyperparameter analysis of ComMem.}
    (A) Effects of learnable modules. 
    (B) Comparison of learning rates for residuals layers ($lr_t$ and $lr_v$) and normalization layers ($lr_n$). 
    (C) Influence of HPC-like memory cache capacity $K$. 
    RN50 and ViT16 denote CLIP-ResNet-50~\citep{resnet} and CLIP-ViT-B/16~\citep{vit} backbones, respectively.
    }
    \label{fig:hyper}
    \vspace{-0.5cm}
\end{figure*}

\paragraph{Effects of Learnable Modules.}
Fig.~\ref{fig:hyper}A presents a progressive ablation of the learnable components in our dual-memory architecture. Starting from the base CLIP-ResNet-50~\citep{radford2021learning} model, we incrementally enable three modules: inter-memory consistency, intra-memory sparsity, and normalization adaptation. Each additional component leads to a consistent improvement across all datasets, demonstrating their complementary contributions.

\paragraph{Learning Rate of Complementary Memory Systems.}  
As shown in Fig.~\ref{fig:hyper}B, we examine the interaction between the learning rates of the residual layers ($lr_v$ for $\delta_v$ and $lr_t$ for $\delta_t$, with $lr_v = lr_t$) and the normalization layers ($lr_n$ for $\theta_n$) using CLIP-ResNet-50~\citep{radford2021learning} backbone. 
The results reveal distinct optimal learning-rate regimes for these two components. 
The normalization layers achieve peak performance at relatively small learning rates (e.g., $10^{-6}\!\sim\!10^{-5}$), aligning with their role in slow and stable integration of visual representation knowledge.
In contrast, the residual layers perform best with larger learning rates (on the order of $10^{-4}$), supporting rapid adaptation to episodic inputs. 
This clear difference in optimal magnitude reflects the complementary dynamics between the neocortical and hippocampal memory systems.

\paragraph{HPC-like Memory Capacity.}
Fig.~\ref{fig:hyper}C analyzes the effect of HPC-like memory capacity $K$ across different datasets and backbones. We observe that performance consistently peaks when $K$ is between 3 and 5. Smaller capacities limit the diversity of stored instances, while larger ones introduce redundancy and noise. This moderate optimal capacity suggests that a compact but selective detailed memory is most effective for balancing flexibility and stability during test-time adaptation.

\subsection{Extended Analysis}
\paragraph{Efficiency Comparison.}
\label{sec:efficiency}

\begin{wraptable}{r}{0.5\textwidth}
\vspace{-1.8cm}
\caption{\textbf{Efficiency comparison on ImageNet~\citep{deng2009imagenet}}. We report the testing time, the achieved accuracy, and the performance gains compared to zero-shot CLIP.}
  \centering
  \resizebox{\linewidth}{!}{
  \begin{tabular}{lcccc}
    \toprule
    Method  & Testing Time & Parameter & Accuracy & Gain \\ 
    \midrule
    CLIP~\citep{radford2021learning}  &  7 min &0.00 M &  59.81&  -\\
    TPT~\citep{shu2022test} &   7 h 33 min & 0.01 M  & 60.74  & +0.93  \\
    DiffTPT~\citep{feng2023diverse} & $>$ 15 h & - &  60.80  & +0.99  \\
    TDA~\citep{karmanov2024efficient} &  53 min & 0.00 M &  61.35  & +1.54  \\
    DPE~\citep{zhang2024dpe} &   1 h 30 min  & 3.91 M  & 63.41 & +3.60 \\
    \textbf{ComMem (Ours)} & 1 h 32 min  & 4.08 M  & \textbf{64.05} & \textbf{+4.24} \\
    \bottomrule
    \end{tabular}
    }
\vspace{-.5cm}
\label{tab:time}
\end{wraptable}

In Table~\ref{tab:time}, we analyze the computational efficiency of ComMem by measuring the total time required to process the 50,000 test samples of ImageNet~\citep{deng2009imagenet}. ComMem achieves its state-of-the-art accuracy in around 1.5 hours. This is substantially more efficient than prompt-tuning methods like TPT~\citep{shu2022test} and DiffTPT~\citep{feng2023diverse} ($>$ 15 hours), which require computationally intensive backpropagation through the text encoder for every sample. While training-free methods like TDA~\citep{karmanov2024efficient} (53 minutes) are faster, ComMem offers a far greater performance gain (a +4.24\% accuracy gain for ComMem vs. +1.54\% for TDA~\citep{karmanov2024efficient}). With only 4.08M learnable parameters, ComMem strikes an excellent balance between high accuracy and practical test-time efficiency.

\begin{wraptable}{r}{0.5\textwidth}
\vspace{-1.cm}
\small
\caption{
\textbf{Comparison of different memory update rules for the HPC-like and NC-like memory systems on the StanfordCars~\citep{krause20133d}.}
Each rule determines how a new feature $\mathbf{f}_v^*$ (for HPC, Sec~\ref{sec:hpc_update}) or a locally optimized prototype $\hat{\mathbf{P}}^t_{\mathrm{local}}$ (for NC, Sec~\ref{sec:nc_update}) is integrated with the existing memory representation (\(\mathbf{p}^v_{\hat{y}}\) or \(\mathbf{P}^t\)). 
Results are reported using CLIP-ResNet-50~\citep{radford2021learning} with a memory capacity of 5 entries per class.
}
\resizebox{\linewidth}{!}{
\begin{tabular}{lccc}
\toprule
\textbf{Update Rule} & \textbf{Formula} & \textbf{HPC Acc.} & \textbf{NC Acc.} \\ 
\midrule
Cumulative Avg. (slow) & $\mathbf{p} \leftarrow \frac{(k-1)\mathbf{p} + \mathbf{f}^*}{k}$ & 58.16 & \textbf{60.19} \\
Cumulative Avg. (fast) & $\mathbf{p} \leftarrow \frac{\mathbf{p} + (k-1)\mathbf{f}^*}{k}$ & 59.71 & 44.31 \\
Exponential Avg. (slow) & $\mathbf{p} \leftarrow 0.99\mathbf{p} + 0.01\mathbf{f}^*$ & 56.62 & 60.08 \\
Exponential Avg. (fast) & $\mathbf{p} \leftarrow 0.01\mathbf{p} + 0.99\mathbf{f}^*$ & 60.07 & 44.55 \\
Prototype Update (slow) & $\mathbf{p} \leftarrow \mathbf{p}$ & 56.75 & 58.28 \\
Full Update (fast) & $\mathbf{p} \leftarrow \mathbf{f}^*$ & \textbf{60.19} & 44.31 \\
\bottomrule
\end{tabular}
}
\vspace{-1.cm}
\label{tab:update}
\end{wraptable}

\paragraph{Different Memory Update Rules.}

As described in Sec.~\ref{sec:hpc_update} and ~\ref{sec:nc_update}, we evaluate different memory update rules between two memory systems. As shown in Table~\ref{tab:update}, the HPC-like detailed memory $\mathbf{P}^v$ achieves optimal performance under \textit{fast update rules}, such as full updates or fast exponential averaging, confirming its role as a rapidly adapting, high-plasticity system. In contrast, the NC-like abstract memory $\mathbf{P}^t$ performs best with \textit{slow update rules} (e.g., cumulative or exponential averaging with small $\lambda_{\text{nc}}$), which promote gradual and stable integration. These complementary findings validate our dual-memory design: the HPC benefits from fast learning for flexible adaptation, whereas the NC evolves slowly to maintain semantic stability.

\paragraph{Visualization of cache features using t-SNE.}

\begin{figure}[h]
    \centering
    \includegraphics[width=0.8\textwidth]{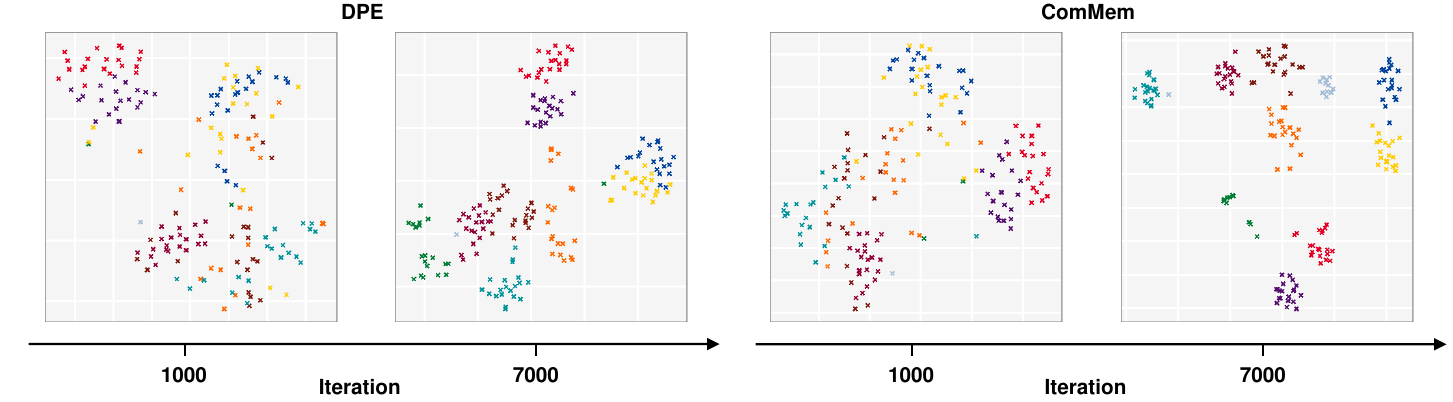}
\caption{t-SNE visualizations of the HPC-like detailed memory (with cache size K = 20) over time using CLIP-ResNet-50~\citep{radford2021learning} on EuroSAT~\citep{helber2019eurosat}.
}
\label{fig:feat_eurosat}
\end{figure}

Fig.~\ref{fig:feat_eurosat} and Appendix Fig.~\ref{fig:feat_food} provides t-SNE visualizations of the HPC-like detailed memory $P^v$ as it evolves. Compared to DPE~\citep{zhang2024dpe}, the feature clusters of ComMem become significantly more compact and more clearly separated by iteration 7000. This indicates that our ComMem successfully learns more discriminative class representations over time, validating our complementary memory mechanism.

\section{Conclusion}
In this paper, we introduced ComMem, a versatile and effective test-time adaptation framework for VLMs inspired by the brain's complementary memory systems. By modeling a fast-adapting HPC-like detailed memory and a slow-consolidating NC-like abstract memory, our approach successfully accumulates and refines multi-modal knowledge from an unlabeled test data stream. The core of our method lies in mimicking systems consolidation and systems reconsolidation, where both memory systems are jointly optimized for each sample to enforce consistency and produce robust predictions. Extensive evaluations on 15 diverse benchmarks demonstrate that ComMem sets a new state-of-the-art, consistently outperforming existing methods in handling natural distribution shifts and generalizing across datasets.

Despite its strong performance, our work has limitations. The use of dynamic memory cache and per-sample optimization slightly increases the computational and memory overhead compared to zero-shot inference, a trade-off for improved accuracy. Future work could explore more efficient implementations of the memory systems, perhaps through sparse or quantized representations. Furthermore, delving deeper into more nuanced brain-inspired mechanisms, such as the process of memory forgetting to prune irrelevant information, could pave the way for even more robust and intelligent adaptation systems.

\paragraph{Acknowledgment}

This work is supported by the Beijing
Major Science and Technology Project (No. Z251100008425003), the STI2030-Major Projects (No. 2022ZD0204900), the NSFC Projects (Nos. 62406160, 92370124, U25B6003, 62350080, 62595773), the Fundamental and Interdisciplinary Disciplines Breakthrough Plan of the Ministry of Education of China (No. JYB2025XDXM101), Beijing Natural Science Foundation (No. L247011), the Shandong Provincial Natural Science Foundation (No. ZR2022ZD01), and the High Performance Computing Center, Tsinghua University.

\clearpage

\newpage






\bibliography{collas2026_conference}
\bibliographystyle{collas2026_conference}

\clearpage
\newpage

\appendix

\setcounter{page}{1}

\section{Related work}\label{sec:related}

\paragraph{Vision-Language Models (VLMs).}
Recent years have witnessed the emergence of powerful VLMs, such as CLIP~\citep{radford2021learning} and ALIGN~\citep{jia2021scaling}, trained on large-scale image-text pairs. These models learn a joint embedding space that aligns visual and textual representations, enabling strong zero-shot generalization across a wide range of downstream tasks~\citep{liang2025comprehensive, dong2025adapting}. To further enhance VLMs' performance, various adaptation strategies have been proposed, most notably in few-shot settings. For example, prompt tuning methods like CoOp~\citep{zhou2022learning} learn continuous prompt vectors to steer model predictions, while adapter-based methods like Tip-Adapter~\citep{zhang2022tip} introduce lightweight modules to adapt features with minimal supervision. While effective, these methods require labeled data, a constraint that our work aims to overcome by operating in a purely test-time setting.

\paragraph{Test Time Adaptation (TTA).}
TTA aims to adapt a model pretrained on a source domain to an unlabeled target domain during inference~\citep{liang2025comprehensive, dong2025adapting}. Early methods primarily focused on updating batch normalization statistics or minimizing prediction entropy to encourage confident outputs~\citep{ioffe2015batch, grandvalet2004semi, yuan2023robust}. Subsequently, Bayesian inference-based methods such as BCA~\citep{zhou2025bca} and BayesTTA~\citep{cui2025bayestta} leverage information from previous test batches for model refinement.
Prompt-based methods such as TPT (Fig.~\ref{fig:summaryTTA}A)~\citep{shu2022test} and its variants~\citep{feng2023diverse, abdul2023align, ma2023swapprompt} learn a tailored prompt for each test sample. However, these methods are inherently amnesic, adapting in isolation without accumulating knowledge over time. More recent efforts, such as HisTPT~\citep{zhang2024historical} and DynaPrompt~\citep{xiao2025dynaprompt} (Fig.~\ref{fig:ComplementaryMemory}B), attempt to address this by dynamically selecting prompts across the test stream, but still operate at the prompt-tuning level, limiting long-term adaptation.

To overcome these limitations, memory-based methods have been introduced.
TDA~\citep{karmanov2024efficient} maintains a dynamic visual cache to enhance adaptation, while DMN-ZS~\citep{zhang2024dual} and DPE~\citep{zhang2024dpe} extend this idea to dual memory systems. DPE evolves visual and textual prototypes jointly, and DMN-ZS incorporates both dynamic test-time and optional static few-shot memories (Fig.~\ref{fig:ComplementaryMemory}C). However, these methods generally treat memory updates as symmetric and fast across modalities, lacking a principled separation of roles. This design leads to overfitting recent examples or under-adapting to novel domains, failing to strike a balance between plasticity (rapid adaptation) and stability (retention of generalizable knowledge).
In contrast, our proposed ComMem framework introduces a brain-inspired architecture that explicitly models distinct fast and slow learning pathways DPE evolves visual and textual prototypes jointly (Fig.~\ref{fig:ComplementaryMemory}D-E and Fig.~\ref{fig:ComMem}). By doing so, it enables both per-sample responsiveness and long-term knowledge accumulation through coordinated cross-modal memory fusion.

\paragraph{Complementary Memory Systems.}
ComMem is grounded in the complementary learning systems (CLS) theory, also referred to as the complementary memory systems, from neuroscience ~\citep{mcclelland1995there, tonegawa2018role}, which explains how the brain learns across multiple timescales. CLS posits a functional specialization between the hippocampus (HPC) that rapidly encodes detailed, episodic experiences and the neocortex (NC) that slowly incorporates structured, generalizable knowledge (Fig.~\ref{fig:ComplementaryMemory}A)~\citep{tonegawa2018role}. The gradual reorganization and transfer of memory traces from HPC to NC is referred to as system consolidation (Fig.~\ref{fig:ComplementaryMemory}B), allowing temporary experiences to become part of stable long-term knowledge~\citep{lei2025reconstructing, golbabaei2025post}. Importantly, consolidated memories can become labile upon retrieval through a process known as memory reconsolidation (Fig.~\ref{fig:ComplementaryMemory}C), which re-engages the HPC to incorporate new information and update existing memories~\citep{lei2025reconstructing}.
To our knowledge, ComMem is the first to explicitly operationalize both the fast (HPC-like) and slow (NC-like) learning dynamics, along with mechanisms analogous to consolidation and reconsolidation, for VLMs' TTA. This biologically grounded design enables ComMem to achieve more robust and generalizable adaptation.


\section{Additional Implementation Details}
\subsection{Experimental Setups}\label{sec:setup}

\textbf{Datasets.}\label{sec:datesets} Our evaluation follows the established benchmarks used in prior work~\citep{shu2022test,feng2023diverse,zhang2024dpe}. To assess \textit{robustness to natural distribution shifts}, we use the ImageNet dataset~\citep{deng2009imagenet} and its four out-of-distribution variants: ImageNet-A~\citep{hendrycks2021natural}, ImageNet-V2~\citep{recht2019imagenet}, ImageNet-R~\citep{hendrycks2021many}, and ImageNet-Sketch~\citep{wang2019learning}. For \textit{cross-dataset generalization}, we evaluate on a diverse suite of 10 recognition datasets: FGVCAircraft~\citep{maji2013fine}, Caltech101~\citep{fei2004learning}, StandfordCars~\citep{krause20133d}, DTD~\citep{cimpoi2014describing}, EuroSAT~\citep{helber2019eurosat},  Flowers102~\citep{nilsback2008automated}, Food101~\citep{bossard2014food}, OxfordPets~\citep{parkhi2012cats}, SUN397~\citep{xiao2010sun}, and UCF101~\citep{soomro2012ucf101}. These datasets offer a comprehensive benchmark for evaluating the robustness of various methods across different distributional variations.

\textbf{Implementation Details.}\label{sec:implement} We employ CLIP~\citep{radford2021learning} models with both ResNet-50~\citep{resnet} and ViT-B/16~\citep{vit} visual backbones for all experiments. Following standard practice, we utilize a broad set of hand-crafted text prompts and also incorporate prompts from CuPL to enhance the initial textual prototypes (Supplementary Table~\ref{tab:prompt}). Followed by previous studies~\citep{karmanov2024efficient,zhang2024dpe}, we generate multiple augmented views for each sample to facilitate robust optimization. The learnable residuals for our complementary memory systems are optimized for a single step per sample using the AdamW optimizer~\citep{loshchilov2017decoupled}. The adaptation of batch normalization or layer normalization parameters~\citep{ioffe2015batch} is performed with a distinct, slower learning rate. Our primary experiments are conducted on a single NVIDIA RTX 3090 GPU with 24GB of memory.

\textbf{Baselines.} We compare ComMem against a comprehensive set of state-of-the-art methods for VLM adaptation. These include: zero-shot \textbf{CLIP}~\citep{radford2021learning} and its prompt \textbf{Ensemble} version; few-shot supervised methods like \textbf{CoOp}~\citep{zhou2022learning}; and leading test-time adaptation methods such as \textbf{TPT}~\citep{shu2022test}, which learns sample-specific prompts; \textbf{DiffTPT}~\citep{feng2023diverse}, which enhances TPT with diffusion-based augmentations; \textbf{DynaPrompt}~\citep{xiao2025dynaprompt}, which performs continual adaptation by adaptively selecting and optimizing relevant prompts for each test sample; \textbf{TDA}~\citep{karmanov2024efficient}, a training-free approach that builds a visual cache; \textbf{TPS}~\citep{sui2025just}, which shifts prototypes at test-time; \textbf{DPE}~\citep{zhang2024dpe}, which evolves dual prototypes, and \textbf{DMN-ZS}~\citep{zhang2024dual}, which uses a dynamic memory for historical test data and a static memory for few-shot data.


\begin{algorithm*}[t]
\caption{\textbf{ComMem: Complementary Memory Systems for Test-Time Adaptation of VLMs}}
\label{alg:commem}
\begin{algorithmic}[1]
\State \textbf{Input}: Pretrained VLM $(\mathcal{E}_v, \mathcal{E}_t)$, target stream $\{\mathbf{x}_t\}_{t=1}^{T}$, 
\Statex \hspace{1.1cm} NC-like abstract memory $\mathbf{P}^t$, HPC-like detailed memory cache $\mathcal{M}$, hyperparameters $(\tau_e, \tau_{\text{conf}}, \lambda_{\text{hpc}}, \lambda_{\text{nc}})$
\State
\For{each incoming test sample $\mathbf{x}_t$}
    \State \textbf{Step 1: Entropy-Guided Visual Encoding} 
    \State Generate $N$ augmented views $\{\mathbf{x}_t^{(i)}\}_{i=1}^{N}$ 
    \For{each view $i=1 \dots N$}
        \State $\mathbf{f}_v^{(i)} \gets \mathcal{E}_v(\mathbf{x}_t^{(i)})$
        \State $p^{(i)} \gets p(y|\mathbf{f}_v^{(i)})$ via Eq.~(\ref{clip_classifier})
        \State $\mathcal{H}^{(i)} \gets -\sum_{c=1}^{C} p_c^{(i)} \log p_c^{(i)}$ \hfill $\triangleright$ Eq.~(\ref{eq:entropy})
    \EndFor
    \State Compute weights $w^{(i)} \propto \exp(-\mathcal{H}^{(i)}/\tau_e)$
    \State Obtain entropy-refined feature $\mathbf{f}_v^* = \sum_i w^{(i)} \mathbf{f}_v^{(i)}$ \hfill $\triangleright$ Eq.~(\ref{eq:entropy_weight})
    \State Assign pseudo-label $\hat{y} = \arg\max p(\cdot|\mathbf{f}_v^*)$
    \State
    \State \textbf{Step 2: HPC Detailed Memory Update} 
    \If{$|\mathcal{M}_{\hat{y}}| < K$}
        \State Add $(\mathbf{f}_v^*, \mathcal{H})$ to $\mathcal{M}_{\hat{y}}$
    \Else
        \State Identify least confident entry $(\mathbf{f}_{\max}, \mathcal{H}_{\max})$
        \If{$\mathcal{H} < \mathcal{H}_{\max}$}
            \State Compute class prototype $\mathbf{p}^v_{\hat{y}} = \text{mean}(\mathcal{M}_{\hat{y}})$
            \State Update $\mathbf{f}'_v = \text{Norm}((1-\lambda_{\text{hpc}})\mathbf{p}^v_{\hat{y}} + \lambda_{\text{hpc}}\mathbf{f}_v^*)$ \hfill $\triangleright$ Eq.~(\ref{eq:hpc_evo})
            \State Replace $(\mathbf{f}_{\max}, \mathcal{H}_{\max}) \leftarrow (\mathbf{f}'_v, \mathcal{H})$
        \EndIf
    \EndIf
    \State Recompute HPC prototypes $\mathbf{P}^v = [\mathbf{p}^v_1,\dots,\mathbf{p}^v_{C_{\text{active}}}]$
    \State
    \State \textbf{Step 3: Test-Time Optimization}
    \State Initialize learnable parameters: residuals $(\delta_t, \delta_v)$, normalization $\theta_n$
    \State Compute updated prototypes:
    \Statex \hspace{1.1cm} $\hat{\mathbf{P}}^t_{\text{local}} = \text{Norm}(\mathbf{P}^t + \delta_t)$ \hfill $\triangleright$ Eq.~(\ref{eq:local_text_proto})
    \Statex \hspace{1.1cm} $\hat{\mathbf{P}}^v = \text{Norm}(\mathbf{P}^v + \delta_v)$
    \State Compute final logits: 
    \Statex \hspace{1.1cm} $\mathbf{z}_{\text{final}} = \mathbf{f}_v^\top \hat{\mathbf{P}}^t_{\text{local}} + \mathcal{A}(\mathbf{f}_v^\top \hat{\mathbf{P}}^v)$
    \State Compute composite loss:
    \Statex \hspace{1.1cm} $\mathcal{L}_{\text{total}} = \mathcal{L}_{\text{ent}} + \mathcal{L}_{\text{align}} + \mathcal{L}_{\text{sparse}}$
    \State Update $(\delta_t, \delta_v, \theta_n)$ with one-step AdamW
    \State Obtain final prediction $\hat{y}_t = \arg\max \text{Softmax}(\mathbf{z}_{\text{final}})$
    \State
    \State \textbf{Step 4: NC Abstract Memory Update (Consolidation)}
    \If{$\mathcal{H}(\mathbf{z}_{\text{final}}) < \tau_{\text{conf}}$}
        \State $\mathbf{P}^t \leftarrow \text{Norm}\big((1-\lambda_{\text{nc}})\mathbf{P}^t + \lambda_{\text{nc}}\hat{\mathbf{P}}^t_{\text{local}}\big)$ \hfill $\triangleright$ Eq.~(\ref{eq:nc_evo})
    \EndIf
\EndFor
\State
\State \Return Adapted VLM $(\mathcal{E}_v, \mathcal{E}_t, \mathbf{P}^t, \mathcal{M})$
\end{algorithmic}
\end{algorithm*}

\subsection{Dataset Details}
\label{subsec:dataset}
In Table~\ref{tab:dataset}, we present the detailed statistics of each dataset we used in our experiments, including the number of classes, the sizes of training, validation and testing sets, and their original tasks.

\begin{table*}[ht]
    \caption{\looseness=-1 \textbf{Detailed statistics of datasets used in experiments}. Note that the last 4 ImageNet variant datasets are designed for evaluation and only contain the test sets.}
    \label{tab:dataset}
    \resizebox{\textwidth}{!}{
    \setlength{\tabcolsep}{2mm}{
    \begin{tabular}{lccccc}
    \toprule
Dataset                  & Classes  & Training & Validation   & Testing & Task \\ \midrule
Caltech101~\citep{fei2004learning} & 100 & 4,128 & 1,649 & 2,465& Object recognition \\
DTD~\citep{cimpoi2014describing}& 47 & 2,820 & 1,128& 1,692 &  Texture recognition\\ 
EuroSAT~\citep{helber2019eurosat}& 10 & 13,500 & 5,400& 8,100 & Satellite image recognition \\ 
FGVCAircraft~\citep{maji2013fine} & 100 & 3,334 & 3,333& 3,333 & Fine-grained aircraft recognition\\
Flowers102~\citep{nilsback2008automated} & 102 & 4,093 & 1,633& 2,463 & Fine-grained flowers recognition \\ 
Food101~\citep{bossard2014food} & 101 & 50,500& 20,200& 30,300 & Fine-grained food recognition  \\ 
ImageNet~\citep{deng2009imagenet} & 1,000 & 1.28M & -& 50,000 & Object recognition \\ 
OxfordPets~\citep{parkhi2012cats} & 37  & 2,944 & 736& 3,669 & Fine-grained pets recognition \\ 
StanfordCars~\citep{krause20133d} & 196 & 6,509 & 1,635& 8,041 & Fine-grained car recognition \\
SUN397~\citep{xiao2010sun}& 397& 15,880 & 3,970& 19,850 & Scene recognition\\ 
UCF101~\citep{soomro2012ucf101}& 101 & 7,639 & 1,898& 3,783 & Action recognition\\
\midrule
ImageNet-V2~\citep{recht2019imagenet} & 1,000 & - & -& 10,000 & Robustness of collocation  \\
ImageNet-Sketch~\citep{wang2019learning} & 1,000 & - & -&50,889 & Robustness of sketch domain\\
ImageNet-A~\citep{hendrycks2021natural}& 200 & - & -&7,500 &Robustness of adversarial attack\\
ImageNet-R~\citep{hendrycks2021many}& 200 & - & -&30,000&Robustness of multi-domains\\
    \bottomrule
    \end{tabular}
    }
    }
\end{table*}

\subsection{Textual Prompts Used in Experiments}
\label{sec:prompt}
In Table~\ref{tab:prompt}, we detail the specific hand-crafted prompts utilized for each dataset. 

\begin{table*}[t]
\centering
    \caption{\looseness=-1 \textbf{Textual prompts used in experiments}. In addition to these prompts, we also employ CuPL~\citep{pratt2023does} prompts to further enhance performance.}
    \label{tab:prompt}
    \resizebox{0.9\textwidth}{!}{
    \setlength{\tabcolsep}{8mm}{
    \begin{tabular}{lc}
    \toprule
Dataset                  & Prompts   \\ \midrule
& ``itap of a \{\texttt{CLASS}\}.'' \\ 
ImageNet~\citep{deng2009imagenet}& ``a bad photo of the \{\texttt{CLASS}\}.'' \\ 
ImageNet-V2~\citep{recht2019imagenet}& ``a origami \{\texttt{CLASS}\}.'' \\ 
ImageNet-Sketch~\citep{wang2019learning}& ``a photo of the large \{\texttt{CLASS}\}.'' \\ 
ImageNet-A~\citep{hendrycks2021natural}& ``a \{\texttt{CLASS}\} in a video game.'' \\ 
ImageNet-R~\citep{hendrycks2021many}& ``art of the \{\texttt{CLASS}\}.'' \\ 
& ``a photo of the small \{\texttt{CLASS}\}.'' \\ \midrule
Caltech101~\citep{fei2004learning} & ``a photo of a \{\texttt{CLASS}\}.'' \\
DTD~\citep{cimpoi2014describing}& ``\{\texttt{CLASS}\} texture.'' \\ 
EuroSAT~\citep{helber2019eurosat}& ``a centered satellite photo of \{\texttt{CLASS}\}.'' \\ 
FGVCAircraft~\citep{maji2013fine} & ``a photo of a \{\texttt{CLASS}\}, a type of aircraft.'' \\
Flowers102~\citep{nilsback2008automated} & ``a photo of a \{\texttt{CLASS}\}, a type of flower.'' \\ 
Food101~\citep{bossard2014food} & ``a photo of \{\texttt{CLASS}\}, a type of food.'' \\ 
OxfordPets~\citep{parkhi2012cats} & ``a photo of a \{\texttt{CLASS}\}, a type of pet.''  \\ 
StanfordCars~\citep{krause20133d} & ``a photo of a \{\texttt{CLASS}\}.'' \\
SUN397~\citep{xiao2010sun}& ``a photo of a \{\texttt{CLASS}\}.''\\ 
UCF101~\citep{soomro2012ucf101}& ``a photo of a person doing \{\texttt{CLASS}\}.'' \\
    \bottomrule
    \end{tabular}
    }
    }
\end{table*}

\subsection{Textual Prompts Used in Experiments}
\label{sec:prompt}
In Table~\ref{tab:prompt}, we detail the specific hand-crafted prompts utilized for each dataset.

\begin{figure*}[t]
    \centering
    \includegraphics[width=0.9\textwidth]{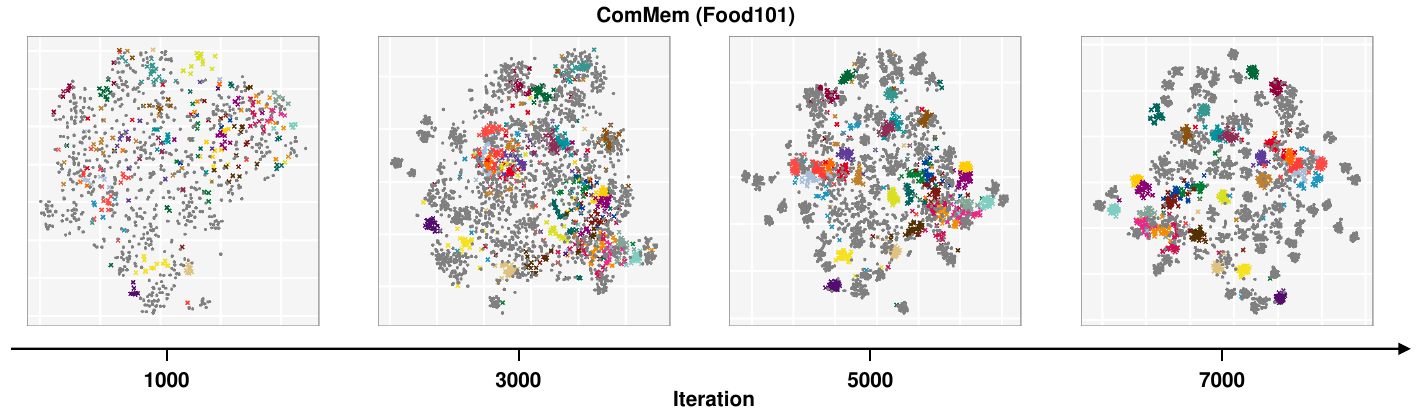}
\caption{t-SNE visualizations of the HPC-like detailed memory (with cache size $K$ = 30) over time using CLIP-ResNet-50~\citep{radford2021learning} on Food101~\citep{bossard2014food}. Features from 10\% of randomly selected classes are shown in different colors, while all others are indicated in gray.
}
\label{fig:feat_food}
\end{figure*}

\end{document}